\documentclass[10pt,twocolumn,letterpaper]{article}

\usepackage{cvpr}
\usepackage{times}
\usepackage{epsfig}
\usepackage{graphicx}
\usepackage{amsmath}
\usepackage{amssymb}
\usepackage{subfigure}
\usepackage{algorithmic}
\usepackage{authblk}
\usepackage[ruled,linesnumbered]{algorithm2e}

\usepackage[breaklinks=true,bookmarks=false]{hyperref}

\cvprfinalcopy 

\def\cvprPaperID{****} 

\begin{document}

\title{\textit{k}-sums: another side of \textit{k}-means}


\author[1]{Wan-Lei Zhao\thanks{Corresponding author: wlzhao@xmu.edu.cn}}
\author[1]{Run-Qing Chen}
\author[1]{Hui Ye}
\author[2]{Chong-Wah Ngo}
\affil[1]{School of Information Science and Engineering, Xiamen University, Xiamen, Fujian, China}
\affil[2]{Department of Computer Science, City University of Hong Kong, Hong Kong}
\renewcommand*{\Affilfont}{\small\it} 
\renewcommand\Authands{ and } 
\date{} 
\maketitle

\begin{abstract}
In this paper, the decades-old clustering method \textit{k}-means is revisited. The original distortion minimization model of  \textit{k}-means is addressed by a pure stochastic minimization procedure. In each step of the iteration, one sample is tentatively reallocated from one cluster to another. It is moved to another cluster as long as the reallocation allows the sample to be closer to the new centroid. This optimization procedure converges faster to a better local minimum over \textit{k}-means and many of its variants. This fundamental modification over the \textit{k}-means loop leads to the redefinition of a family of \textit{k}-means variants. Moreover, a new target function that minimizes the summation of pairwise distances within clusters is presented. We show that it could be solved under the same stochastic optimization procedure. This minimization procedure built upon two minimization models outperforms \textit{k}-means and its variants considerably with different settings and on different datasets.
\end{abstract}

\section{Introduction}
\label{sec:intro}
Clustering is a basic processing tool in many areas such as data mining~\cite{ml04:zhao}, data compression~\cite{JPDSPS11},  pattern recognition and computer vision. Since the first \textit{k}-means methods~\cite{km82, kmeans} was proposed in year \textit{1982}, various clustering methods~\cite{Jain99} have been proposed one after another in the last three decades. These methods range from classic density based methods such as mean shift~\cite{meansift}, DB-SCAN~\cite{dbscan}, and recent clusterDP~\cite{science14}, to graph based methods such as spectral clustering~\cite{spectral} and Rank-Order~\cite{rankorder}, etc. Nevertheless, \textit{k}-means~\cite{kmeans} remains popular for its efficiency, versatility as well as simplicity. According to~\cite{top10}, it is recognized among the top ten most popular methods in data mining. 

Given $n$ data samples in \textit{d}-dimensional space $\mathbb R^d$, and an integer \textit{k}, the clustering task is modeled as a distortion minimization process in \textit{k}-means. In one iteration, it assigns $n$ samples to one of \textit{k} sets where its corresponding centroid is the closest to the sample. The minimization target function is given as
\begin{equation}
        \mbox{Min. }\sum_{q(x_i)=r}d(x_i, C_r),
        \label{eqn:tkm}
\end{equation}
where $x_i \in \mathbb R^d$ and $C_r$ is the centroid of cluster $r$. In Eqn.~\ref{eqn:tkm}, function $q(x_i)$ returns the closest centroid (among \textit{k} centroids) for sample $x_i$. There are in general three major steps involved in \textit{k}-means iterations. In the initial step, \textit{k} samples are randomly selected as the initial centroids. In the assignment step, each sample is assigned to its closest centroid. In the centroid updating step, each centroid $C_r$ is updated by taking the average over the assigned samples. The last two steps are repeated until there is no distortion variation (Eqn.~\ref{eqn:tkm}) in the two consecutive iterations. This iteration process is widely known as the classic ``egg-chicken'' loop. 

Although it is simple and effective, the major issues for this ``egg-chicken'' loop are in several aspects. Firstly, the target function is minimized in an implicit manner. The iteration in its nature minimizes the discrepancy between two consecutive iterations instead of Eqn.~\ref{eqn:tkm}. Moreover, the update on the centroid is postponed to the moment when all the samples are assigned to their closest centroids. Given $t$ and $t+1$ are two consecutive iterations in \textit{k}-means, the real target function that is minimized during the iteration is
\begin{equation}
        \mbox{Min. }\sum_{q(x_i)=r}d(x_i, C^{(t)}_r).
        \label{eqn:tkm1}
\end{equation}
After the assignment step, samples assigned to $C^{(t)}_r$ are averaged to produce $C^{(t+1)}_r$. Such kind of minimization is inefficient in the sense that the samples are compared to centroids produced from previous iteration \textit{t}. No update happens when a sample is moved from other clusters to \textit{r}. However, according to Eqn.~\ref{eqn:tkm}, the centroids are expected to be updated as soon as the membership of one sample changes. Due to the delayed update, the samples are not allowed to compare with the centroids that reflect the real structure of clusters at each moment. For the above reasons, usually \textit{k}-means converges slowly to a local minimum.

In the literature, several efforts have been devoted to enhancing the clustering quality. Particularly, the clustering quality is boosted by a careful seeding scheme~\cite{kpp07, ikm}, for which the centroids are initialized based upon the data distribution. Recently, the \textit{k}-means  problem is approximated by a maximization procedure~\cite{boostkmeans,ml04:zhao}. Encouraging performance is achieved.

In this paper, the \textit{k}-means clustering that is formulated in Eqn.~\ref{eqn:tkm} is addressed by an explicit stochastic minimization process. It turns out to be simpler as well as better over \textit{k}-means and many of its variants. Under the same minimization framework, a family of \textit{k}-means variants such as hierarchical \textit{k}-means and Sequential \textit{k}-means is redefined to achieve better performance. Moreover, a new target function that minimizes the summation of pairwise distances within each cluster is proposed. Based on the same stochastic optimization procedure, the target function is explicitly minimized with the same time complexity as the conventional \textit{k}-means.

The remainder of this paper is organized as follows. The reviews on the representative \textit{k}-means variants are presented in Section~\ref{sec:relate}. In Section~\ref{sec:methd}, the driven function derived from the \textit{k}-means target function is presented. In addition, a new clustering target function and its driven function are proposed. The iteration procedures built upon these two driven functions are accordingly presented. The possible extensions, convergence and complexity analysis are presented in Section~\ref{sec:ext}. The effectiveness of the proposed methods is studied in Section~\ref{sec:exp}. Section~\ref{sec:conc} concludes the paper.

\section{Related Work}
\label{sec:relate}
\textit{k}-means has been widely adopted as a basic tool in data mining~\cite{ml04:zhao}, various data preprocessing and pattern recognition~\cite{Jain99} mainly due to its versatility and simplicity. Various improvement schemes are proposed during the last three decades to boost its performance in terms of either clustering quality or scalability.

A representative work in improving the clustering quality was proposed by S. Vassilvitskii et al.~\cite{kpp07,kpp12}. In the method, the initial centroids are selected to be far from each other to reflect the underlying data distribution. It leads to higher clustering quality as well as faster convergence speed according to~\cite{kpp07}. However, \textit{k-1} rounds of scanning over the whole data are necessary to find the initial centroids. The number of scanning rounds has been successfully reduced to a few~\cite{kpp12} or even fewer~\cite{nips16bachem}. However, all the above improvements focus on the initial assignment stage. The ``egg-chicken'' loop is still adopted. Therefore the aforementioned pitfalls that are caused by this loop remain unchanged.

In the literature, there are several efforts aiming to transform the ``egg-chicken'' loop into an optimization procedure~\cite{ml04:zhao,mlpr10:matus,ijcai13:noam,boostkmeans}. In~\cite{ml04:zhao}, \textit{k}-means is addressed as a maximization problem under \textit{Cosine} distance. This maximization solution is extended to the whole $l_2$-space in~\cite{boostkmeans}. While following  Hartigan procedure~\cite{cluster75:hartigan}, methods from~\cite{mlpr10:matus,ijcai13:noam} perform the distortion minimization directly on the original \textit{k}-means target function. There are two major differences in these methods from the other \textit{k}-means variants. Firstly,  a cluster and its corresponding centroid are updated as soon as the membership of one sample changes during the iteration. Secondly, the target function in each update step is monotonically optimized in a greedy manner. Another interesting discovery from~\cite{boostkmeans} is that the improvement achieved from careful seeding~\cite{kpp07} is minor in comparison to that from the modification of the iteration procedure. Nevertheless, the maximization model in~\cite{ml04:zhao} only works under \textit{Cosine} distance. Although the methods in~\cite{mlpr10:matus,ijcai13:noam,boostkmeans} are feasible in the whole $\textit{l}_2$-space, the optimization converges in a slow pace as it has to guarantee a monotonic optimization in each update step. Specifically, a sample is not necessarily assigned to its closest centroid in one update~\cite{mlpr10:matus,ijcai13:noam}, which actually hinders the optimization process from reaching a better local optimum.

Although the time complexity of \textit{k}-means is linear to the size of the dataset, it could become very slow as both \textit{k} and dataset size \textit{n} are large. The processing bottleneck comes from the operations of assigning samples to their closest centroids in every \textit{k}-means iteration. As a result, many efforts have been made to speed-up the sample-to-centroid comparison. Solutions presented in~\cite{wsdm14,pelleg99} reduce the comparisons with the support of indexing structures such as inverted file or KD-tree. However, the former is only effective for sparse vectors, while the latter performs poorly on dense high-dimensional vectors. The scalability issue of \textit{k}-means is also addressed by subsampling strategy. In methods such as Mini-Batch~\cite{mnkm10} and \cite{icdm04}, only a small portion of the whole dataset are sampled to update the cluster centroids. Such methods usually achieve high speed efficiency at the expense of low clustering quality.

Besides aforementioned \textit{k}-means variants, there are still another two popular variants, namely hierarchical \textit{k}-means~\cite{jain88} and Sequential \textit{k}-means~\cite{mac67}. Hierarchical \textit{k}-means conducts the clustering in a top-down hierarchical manner~\cite{jain88,ml04:zhao,kddzhao05}. The clustering solution is obtained via a sequence of repeated partitions over intermediate clusters. When the fanout on each hierarchy is \textit{2}, it is called as ``bisecting \textit{k}-means''~\cite{ml04:zhao}. The advantages of such scheme are two folds. Firstly, it is able to produce a dendrogram view of the dataset. Moreover, the clustering time complexity of \textit{k}-means is reduced from $O(t{\cdot}k{\cdot}n{\cdot}d)$ to $O(t{\cdot}log(k){\cdot}n{\cdot}d)$~\cite{boostkmeans}, where \textit{t} is the number of iterations. This is significant when \textit{n}, \textit{d}, and \textit{k} are all very large. The dark side is that clustering performance could be poor as it breaks  \textit{Lloyd}'s condition~\cite{boostkmeans}. Sequential \textit{k}-means is also known as online \textit{k}-means. It is designed for the case that samples come in sequentially. The clustering centroid is updated incrementally as a new sample joins in~\cite{mac67}. Given $C_r$ is the closest centroid to sample $x_i$ and $n_r$ is the size of cluster \textit{r}, the centroid is updated by
\begin{equation}
   C_r=C_r+\frac{x_i-C_r}{n_r+1}.
\label{eqn:seqkm}
\end{equation}
Different from the conventional \textit{k}-means, it is supposed that there is only one single pass over the data, although it can be trivially repeated multiple times to reallocate samples until convergence. 

Overall, although the various modifications are made over conventional \textit{k}-means in the literature, most of the variants still build upon the ``egg-chicken'' loop. In this paper, the modification is undertaken on the ``egg-chicken'' loop itself. This leads to a fundamental change over \textit{k}-means. It becomes simpler and considerably better while involving no additional computational costs. More importantly, this new iteration procedure can be easily implanted in various \textit{k}-means variants to boost their performance.

\section{\textit{k}-sums Clustering}
\label{sec:methd}
As discussed in Section~\ref{sec:intro}, the major issues that lie in the conventional \textit{k}-means loop are that the centroids are not updated timely and the target function is not explicitly minimized. In the following, we are going to show it is possible to minimize Eqn.~\ref{eqn:tkm} directly by a stochastic optimization procedure. The optimization is driven by a function that minimizes Eqn.~\ref{eqn:tkm} greedily. This function is called as driven function $\mathcal{I}_m$. In addition, another target function that aims to minimize the summation of pairwise distances within each cluster is presented. Similarly, a driven function given as $\mathcal{I}_s$ is derived for this target function. We show that both minimization problems could be solved by the same stochastic optimization procedure. 

To facilitate our discussions in this section and the later, several variables are introduced. The \textit{k} clusters produced by a clustering method are given as $\{S_1,\cdots,S_r,\cdots,S_k\}$. Accordingly, the sizes of the clusters are given as $n_1,\cdots,n_r,\cdots,n_k$. The composite vector of one cluster is defined as $D_r=\sum_{x_i \in S_r}x_i$\footnote{Both $x_i$ and $D_r$ are column vectors by default.}, which is nothing more than the summation of the samples in one cluster. The cluster centroid $C_r$ is given as $C_r=\frac{D_r}{n_r}$.

In the following, we are going to first show the driven functions for two optimization problems. Based on the driven functions, the novel \textit{k}-means iteration procedure is presented. 

\subsection{Driven Function $\mathcal{I}_m$}
Given a sample $x_i$, it is currently located in cluster $S_w$, namely $x_i \in S_w$. According to Eqn.~\ref{eqn:tkm}, its distance to the centroid of $S_w$ is given as
\begin{equation}
d(x_i, C_w)=\parallel x_i - \frac{D_w}{n_w}\parallel^2.
\label{eqn:d2c}
\end{equation}
This is also the distortion associated with sample $x_i$ that contributes to Eqn.~\ref{eqn:tkm}.

Let's now assume that the structure of cluster $S_w$ has been changed in the previous iterations as some samples have been swapped in/out. For this reason, $C_w$ may be no longer the closest centroid for $x_i$. Now we check whether there exists any other cluster $S_v$ ($v \neq w$) that is more appropriate for $x_i$. The distance between $x_i$ and $C_v$ is measured supposing that $x_i$ is already joined into cluster $S_v$. As a result, the distortion variation for $x_i$ is given as Eqn.~\ref{eqn:deltai1} for this possible movement.
\begin{equation}
	\label{eqn:deltai1}
	\begin{aligned}
	\mathcal{I}_m(x_i, w, v) = d(x_i, C_w) - d(x_i, C_v)&,\\
	\text{where } C_v=\frac{D_v+x_i}{n_v+1}&.
	\end{aligned}
\end{equation}
Please be noted that Eqn.~\ref{eqn:deltai1} is different from online \textit{k}-means~\cite{mac67} in the sense that $x_i$ is supposed to be a member of cluster $S_v$, rather than excluding $x_i$ out from $S_v$ in the distance evaluation. In above equation, as $\mathcal{I}_m(x_i, w, v) > 0$, assigning $x_i$ to cluster $S_v$ will decrease the distortion associated with $x_i$, which in turn leads to the possible decrease in the overall distortion for target function Eqn.~\ref{eqn:tkm}. So the sample is moved from the current cluster to $S_v$ as long as $\mathcal{I}_m(x_i, w, v)$ is positive and the maximum among all $k-1$ tentative re-allocations. The movement of sample $x_i$ from cluster $S_w$ to $S_v$ involves the update of membership for $x_i$ as well as the update on $C_v$, $n_v$, $C_w$ and $n_w$. Function $\mathcal{I}_m(x_i, w, v)$ is therefore called as driven function. 

This driven function is essentially different from~\cite{mlpr10:matus,ijcai13:noam}, in which the distance between $x_i$ and $C_w$ is calculated assuming $x_i$ has been removed out from $S_w$. This subtle difference leads to the very different interpretations about the effect. Function $\mathcal{I}_m(x_i, w, v)$ guarantees that $x_i$ is placed to its closest centroid. While there is no guarantee that the movement of $x_i$ leads to the decrease in Eqn.~\ref{eqn:tkm}. The function in~\cite{mlpr10:matus,ijcai13:noam} leads to the opposite effects. Namely, the movement of $x_i$ leads to the lower of overall distortion in Eqn.~\ref{eqn:tkm}, however $x_i$ is not necessarily put into the cluster that is closest to it. In other words, $\mathcal{I}_m(x_i, w, v)$ allows the ``individual interests'' to be maximized in each movement, while function in~\cite{mlpr10:matus,ijcai13:noam} guarantees the monotonic increase of ``general interests'' in each movement. As analyzed in Section~\ref{sec:cmplx}, the former is less likely being trapped in a local optimum and therefore performs  considerably better as is revealed in the experiments.

To simplify the computation, the distance between $x_i$ and $C_w$ is given as
\begin{equation}
d(x_i, C_w)=\frac{\parallel{n_w{\cdot}x_i-D_w}\parallel^2}{n_w^2}.
\label{eqn:x2cw}
\end{equation}
Accordingly, the distance between $x_i$ and $C_v$ is given as

\begin{equation}
d(x_i, C_v)=\frac{\parallel{n_v{\cdot}x_i-D_v}\parallel^2}{(n_v+1)^2}.
\label{eqn:x2cv}
\end{equation}
%

In some scenarios, we may use \textit{Cosine} distance instead of $\textit{l}_2$-norm to measure the distance between samples and the distance between samples and the centroids. One would have the following equations to measure the distance between sample $x_i$ and centroid $C_w$ and $C_v$ respectively.

\begin{equation}
cos(x_i, C_w)=\frac{x_i'{\cdot}D_w}{\sqrt{x_i'{\cdot}x_i}{\cdot}\sqrt{D_w'{\cdot}D_w}}
\label{eqn:cosx2cw}
\end{equation}

\begin{equation}
cos(x_i, C_v)=\frac{x_i'{\cdot}D_v+x_i'{\cdot}x_i}{\sqrt{x_i'{\cdot}x_i}{\cdot}\sqrt{D_v'{\cdot}D_v+2x_i'{\cdot}D_v+x_i'{\cdot}x_i}}
\label{eqn:cosx2cv}
\end{equation}
Since the $\textit{l}_2$-norm of $x_i$ could be pre-computed, the terms we should consider in Eqn.~\ref{eqn:cosx2cw} and Eqn.~\ref{eqn:cosx2cv} are the inner-products between $x_i$ and the composite vectors, and the $\textit{l}_2$-norms of composite vectors $D_w$ and $D_v$.

It is clear to see that $C_w$ and $C_v$ are not involved in any case of the distance computation. Only $D_r$s and $n_r$s are required. The composite vectors $D_r$s are nothing more than \textit{k} summations of samples within \textit{k} clusters. To this end, the ``means'' are replaced by ``sums''. For this reason, our new clustering method is called as \textbf{\textit{k}-sums} from now on. Please be noted that it is possible to formulate the driven function (Eqn.~\ref{eqn:deltai1}) in terms of centroids. However, the computing cost of updating centroids turns out to be much higher than updating only the composite vectors as the update operation is frequently undertaken in the iteration.

\subsection{Driven Function $\mathcal{I}_s$}
In some scenarios, defining the centroid for a clustering problem would be hard or even impossible. For instance, the sample vectors could not be averaged when the values in each data dimension/property are discrete. A good case is the gender of a person. This is where the clustering method such as PAM~\cite{pam87:kaufman} comes, in which cluster modes instead of centroids are defined. Moreover, the criterion of being a cluster may change. Instead of minimizing summations of distances to a mode/centroid, we may need to minimize the intra-distances within each cluster. This leads to a new target function. Namely, the target function is simply defined as

\begin{equation}
\mbox{Min. } \sum_{r=1}^k\sum_{i,j \in S_r \& i < j}d(x_i, x_j).
\label{eqn:ksumi2}
\end{equation}
Notice that this minimization target function is different from $\mathcal{I}_2$ proposed in~\cite{ml04:zhao}, because it aims to minimize the weighted intra-distances within each cluster. In~\cite{ml04:zhao}, the average pairwise distance within each cluster is weighted by the size of a cluster. To seek for the optimal solution for Eqn.~\ref{eqn:ksumi2}, intuitively one has to try out all the possible combinations of the samples in one cluster. This is unfortunately NP-hard as PAM~\cite{pam87:kaufman}. As a consequence, we only seek for a local minimal solution to this problem. In particular, in $\textit{l}_2$-space, this target function can be addressed with a greedy procedure in a very efficient fashion. 

Given that $x_i \in S_w$ and the distance between samples is measured by $\textit{l}_2$-norm, the overall distance between sample $x_i$ and cluster $S_w$ is defined as

\begin{equation}
d(x_i, S_w)=\sum_{x_j \in S_w}\parallel x_i - x_j \parallel^2,
\label{eqn:x2Sr}
\end{equation}
which is the summation of distances between sample $x_i$ and each sample in $S_w$. Eqn.~\ref{eqn:x2Sr} can be further simplified as

\begin{equation}
d(x_i, S_w)=n_w{\cdot}x_i'{\cdot}x_i-2{\cdot}x_i'{\cdot}D_w+\sum_{x_j \in S_w}{x_j'{\cdot}x_j},
\label{eqn:x2Srv2}
\end{equation}
where $D_w$ is the composite vector of cluster $S_w$. Eqn.~\ref{eqn:x2Srv2} can be efficiently calculated given the $\textit{l}_2$-norms of each sample can be pre-computed and kept in a look-up table. The second term is the inner-product between sample $x_i$ and the composite vector, which is comparable to calculating the distance between sample $x_i$ and a centroid in the conventional \textit{k}-means model. Given sample vectors are $\textit{l}_2$-normalized, Eqn.~\ref{eqn:x2Srv2} is further simplified as

\begin{equation}
d(x_i, S_w)=2{\cdot}n_w-2{\cdot}x_i'{\cdot}D_w.
\label{eqn:x2Srcos}
\end{equation}
Eqn.~\ref{eqn:x2Srcos} could be used as \textit{Cosine} distance when we want to adopt \textit{Cosine} to measure the distances between vectors. They are interchangable as the vectors are $\textit{l}_2$-normalized. 

Now let's consider the similar driven strategy that we derive for target function Eqn.~\ref{eqn:tkm}. Given sample $x_i \in S_w$, we consider whether it could be better if we put $x_i$ into $S_v$. The distance between $x_i$ and $S_v$ is given as
\begin{equation}
	\label{eqn:x2Sv}
	\begin{aligned}
		d(x_i,S_v)=(n_v&+1){\cdot}x_i'{\cdot}x_i-2{\cdot}x_i'{\cdot}(D_v+x_i)\\
		&+\sum_{x_j \in S_v \& j \neq i}{x_j'{\cdot}x_j} +x_i'{\cdot}x_i\\
		=n_v{\cdot}&x_i'{\cdot}x_i-2{\cdot}x_i'{\cdot}D_v+\sum_{x_j \in S_v \& j \neq i}{x_j'{\cdot}x_j}.\\
	\end{aligned}
\end{equation}

Comparing distance $d(x_i, S_v)$ to $d(x_i, S_w)$, it is easy to judge whether such movement is ``profitable'' for $x_i$. Namely, we work out the driven function to minimize target function Eqn.~\ref{eqn:ksumi2} as
\begin{equation}
\mathcal{I}_s(x_i, w, v)= d(x_i, S_w) - d(x_i, S_v).
\label{eqn:deltai2}
\end{equation}
As shown in Eqn.~\ref{eqn:x2Srv2} and Eqn.~\ref{eqn:x2Sv}, it is unnecessary to maintain $C_r$. Similar as driven function $\mathcal{I}_m$, one only needs to maintain $D_r$s and $n_r$s during the optimization for computational efficiency. 

In the minimization step, we check Eqn.~\ref{eqn:deltai2} with all $k-1$ clusters, and move $x_i$ to the cluster where $\mathcal{I}_s(x_i, w,v)$ is positive and the maximum. Notice that each such kind of movement will lead to a steady decrease in the target function (Eqn.~\ref{eqn:ksumi2}). While it is not guaranteed that Eqn.~\ref{eqn:tkm} steadily decreases when driven by $\mathcal{I}_m$.

\begin{figure*}[t]
\begin{center}
	\subfigure[Measured with Eqn.~\ref{eqn:tkm} driven by $\mathcal{I}_m$]
	{\includegraphics[width=0.255\linewidth]{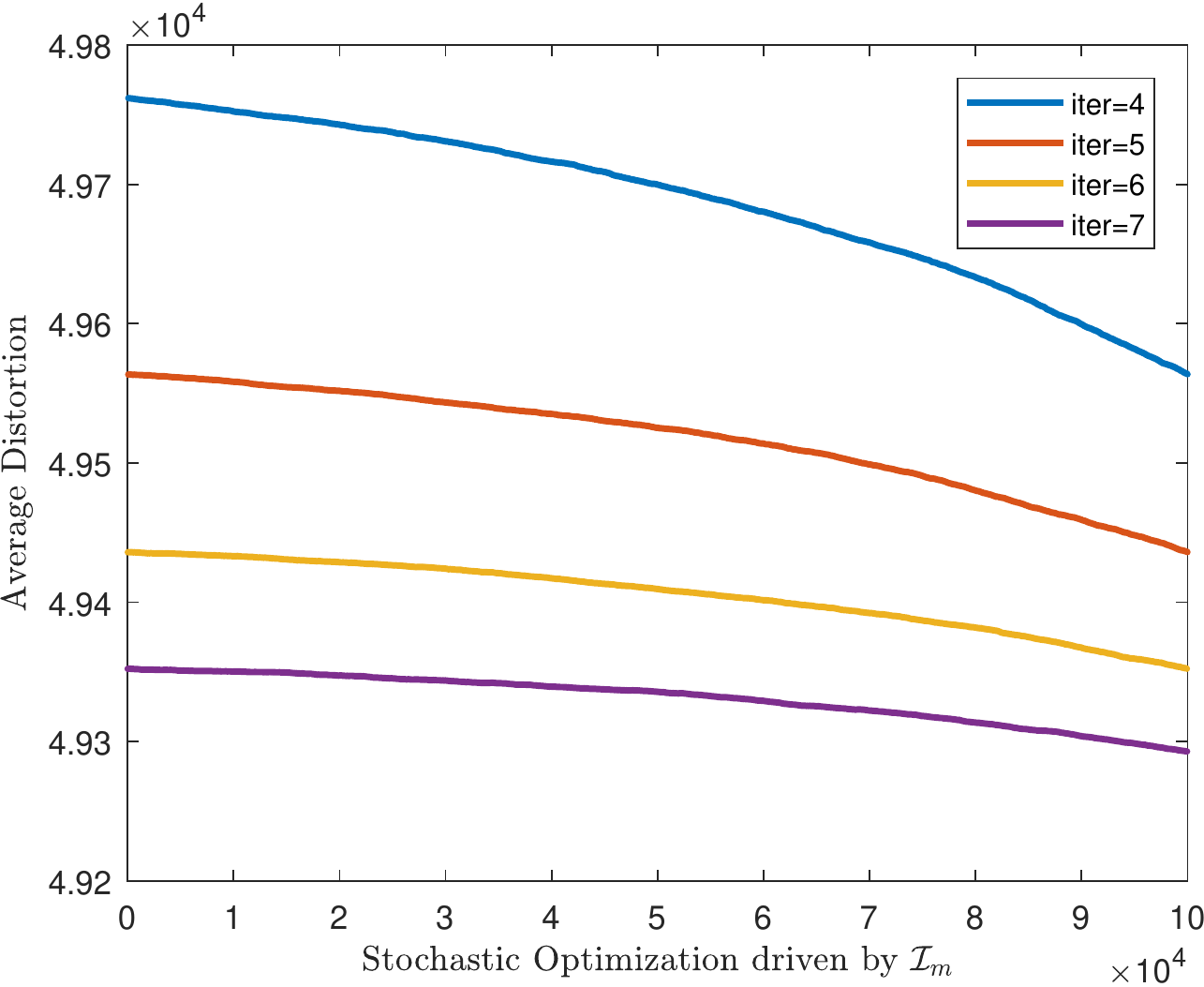}}
	\hspace{0.01in}
	\subfigure[Zoom-in view on the curve of Iter=4 in figure (a)]
	{\includegraphics[width=0.20\linewidth]{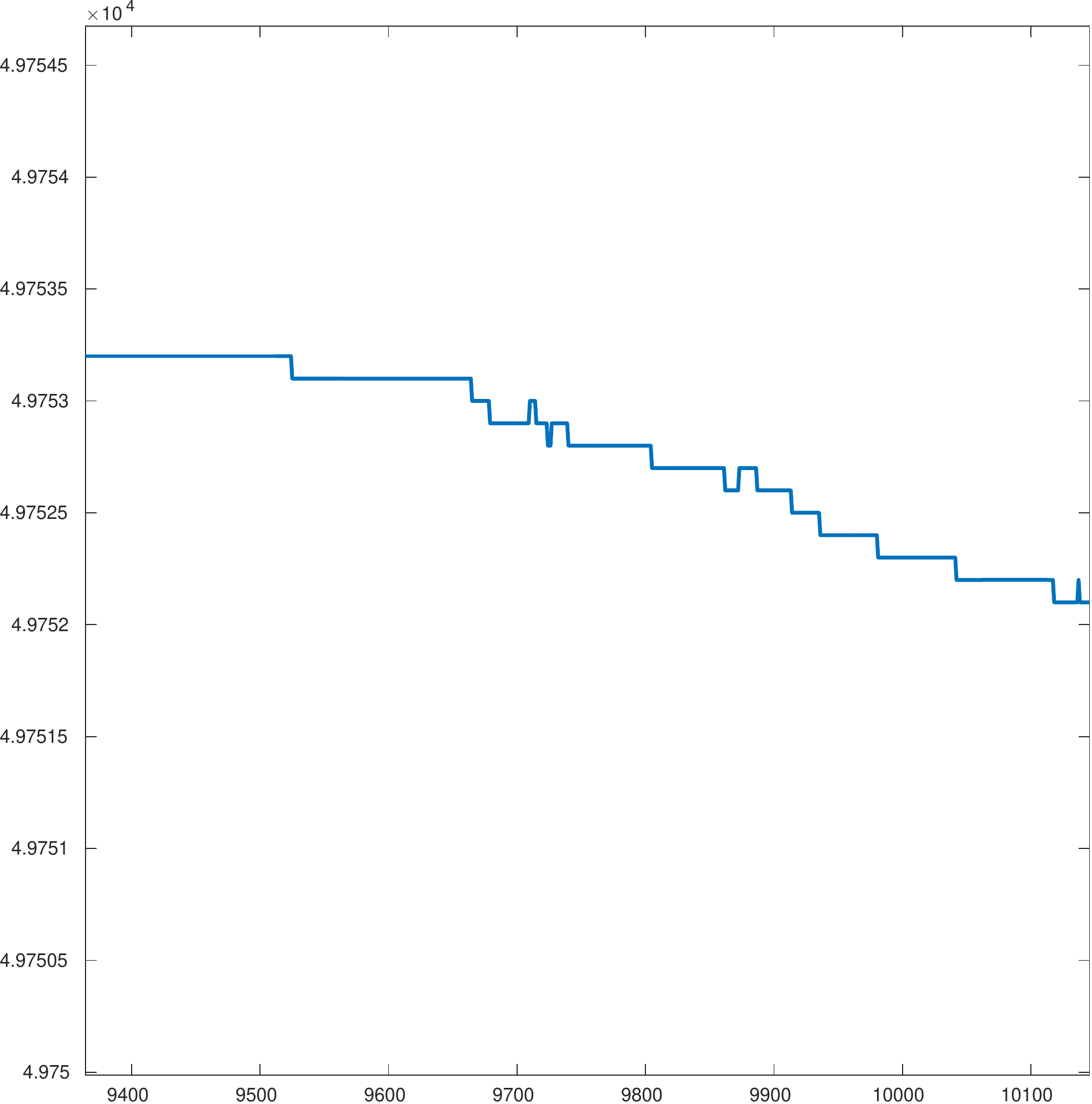}}
	\hspace{0.01in}
	\subfigure[Measured with  Eqn.~\ref{eqn:ksumi2} driven by $\mathcal{I}_s$]
	{\includegraphics[width=0.255\linewidth]{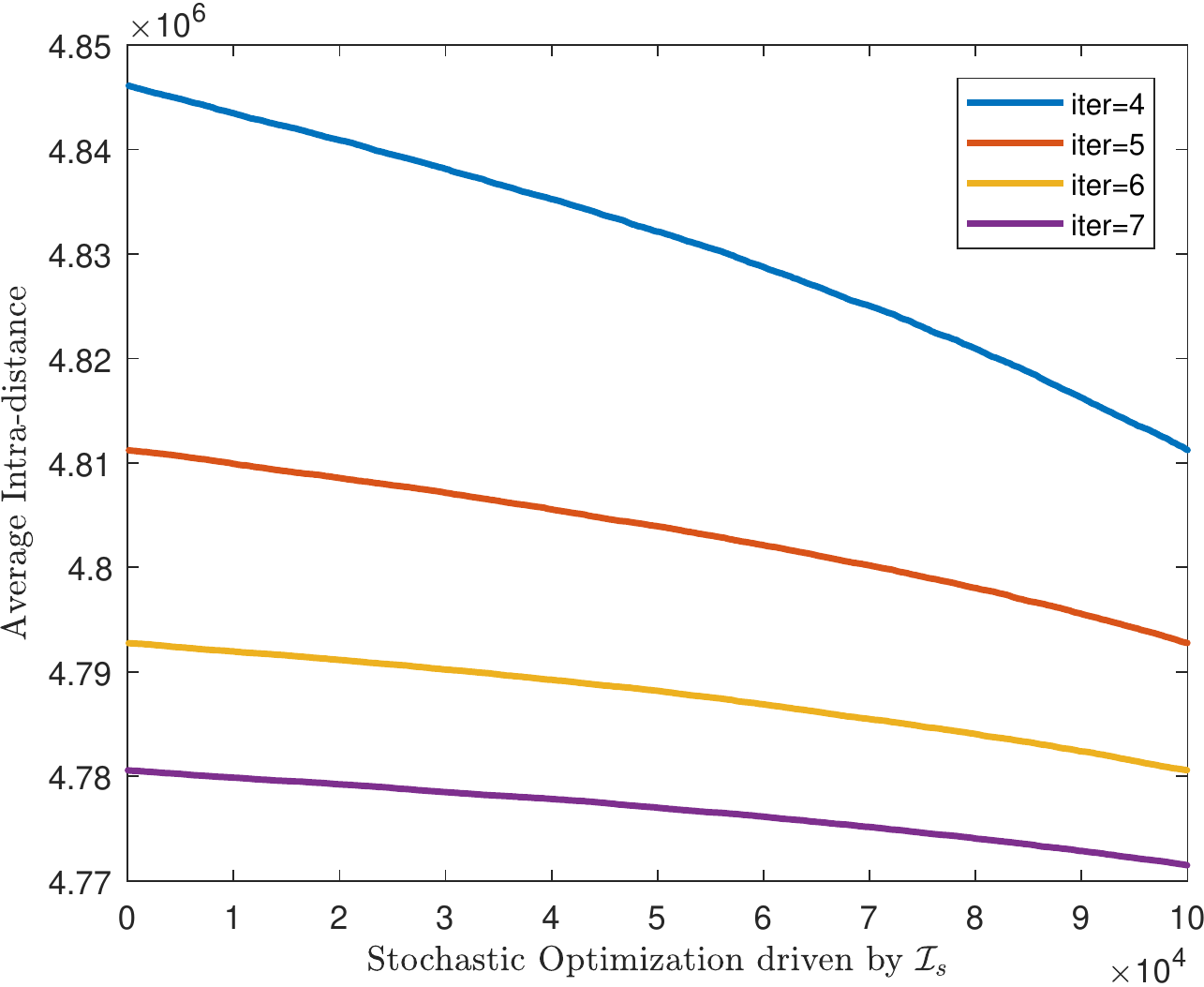}}
	\hspace{0.01in}
	\subfigure[Measured with Eqn.~\ref{eqn:tkm} driven by $\mathcal{I}_s$]
	{\includegraphics[width=0.255\linewidth]{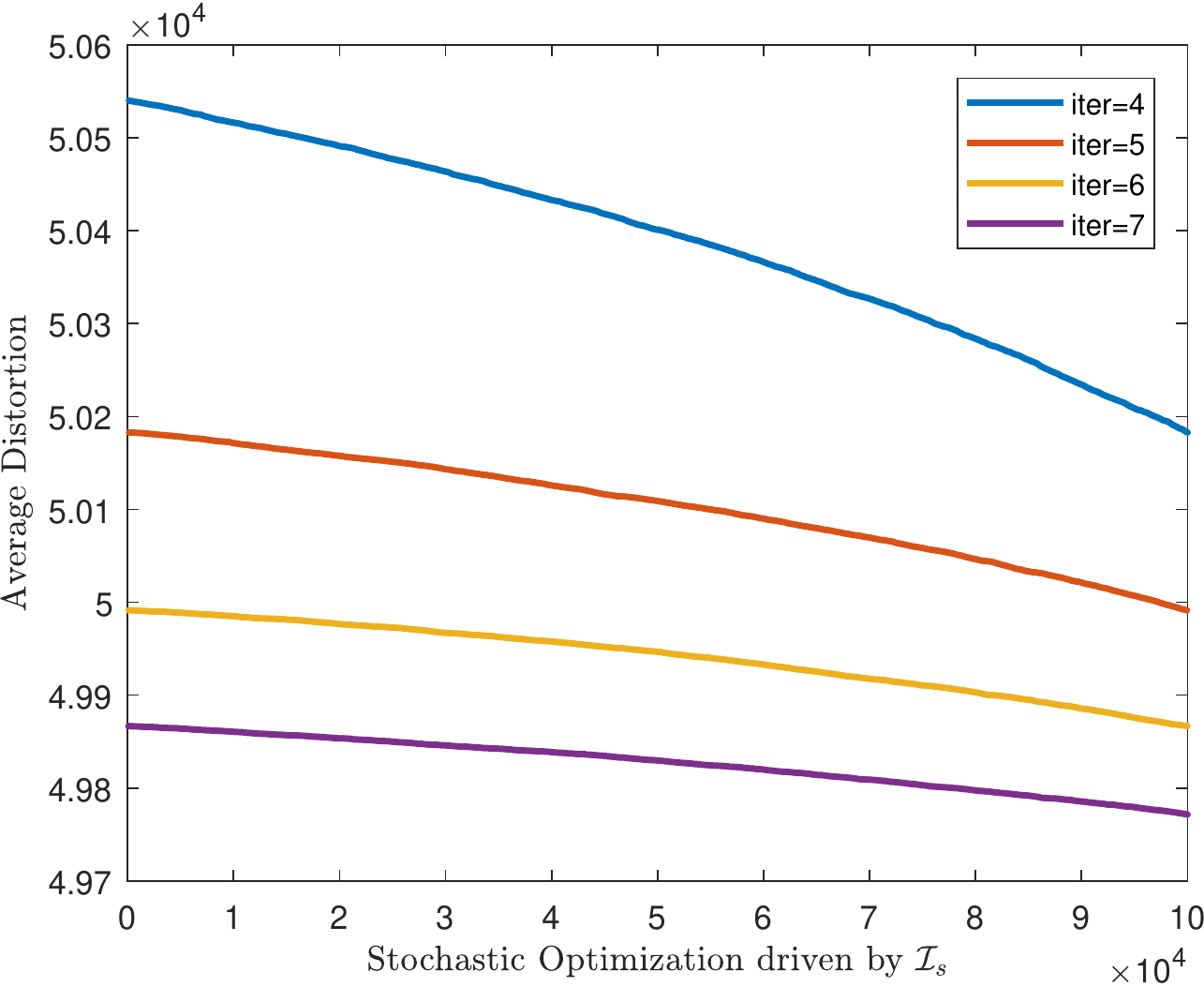}}
\end{center}
\caption{The target function variation curves produced on \textit{100}K SIFT data by Alg.~\ref{alg:ksumi1} on four consecutive iterations ($4{\sim}7$). Alg.~\ref{alg:ksumi1} is driven by $\mathcal{I}_m$ (figure~(a)) and $\mathcal{I}_s$ (figure~(c)) respectively. The function value measured by Eqn.~\ref{eqn:tkm} when Alg.~\ref{alg:ksumi1} is driven by $\mathcal{I}_s$ is shown in figure~(d). All function values are normalized by the size of dataset.}
\label{fig:dist4iter}
\end{figure*}

\subsection{Stochastic Optimization Procedure}
With two driven functions $\mathcal{I}_m$ and $\mathcal{I}_s$  derived in the above sections, it becomes natural to work out the clustering iteration. Since the optimization procedures for $\mathcal{I}_m$ and $\mathcal{I}_s$ are similar, let's take $\mathcal{I}_m$ as an example. In one step of the iteration, sample $x_i$ is randomly selected, then it is checked with $k-1$ clusters to seek for the maximal $\mathcal{I}_m$. A sample reallocation is undertaken as long as $\mathcal{I}_m$ reaches the maximum and is positive. The details of the clustering method \textit{k}-sums are presented in Alg.~\ref{alg:ksumi1}, which is in general similar as~\cite{mlpr10:matus,boostkmeans} yet driven by different function.

\begin{algorithm}
	\KwData{Input: $X_{d{\times}n}$, $k$}
	\KwResult{Output: $S_1,{\cdots},S_r,{\cdots},S_k$}
	Lables[$1, \cdots, n$]$\leftarrow 0$\;
	Assign each $x_i \in X$ with a random cluster label\;
	Calculate $D_1,{\cdots},D_r,{\cdots},D_k$ and $n_1,{\cdots},n_r,{\cdots},n_k$\;
	\While{{not convergent}}{
		\For{each $x_i \in X$ (in random order)}{
			$w \leftarrow$Labels[i]\;
			Seek $S_v$ that $\mathcal{I}_m(x_i, w, v)$ reaches the maximum\;
			\If {$\mathcal{I}_m(x_i, w, v) > 0$}{
				Lables[$i$]$\leftarrow v$\;
				$D_w \leftarrow D_w - x_i$; $n_w \leftarrow n_w - 1$\;
				$D_v \leftarrow D_v + x_i$; $n_v \leftarrow n_v + 1$\;
			}
		}
	}
	\caption{\textit{k}-sums driven by $\mathcal{I}_m$}
	\label{alg:ksumi1}
\end{algorithm}

As shown in Alg.~\ref{alg:ksumi1}, following the practice in~\cite{boostkmeans}, no initial centroid selection or initial sample-to-centroid assignment is involved in \textit{k}-sums. Each sample is assigned with a random cluster label. With these random labels, it is possible to calculate $D_r$s and $n_r$s (Alg.~\ref{alg:ksumi1}, \textit{Line 3}). At the beginning, the samples from different clusters are mixed up with each other at the initial stages~\cite{boostkmeans}. However, the boundaries between clusters become clearer after only a few iterations. In each iteration, samples are evaluated in random order with $\mathcal{I}_m$. The movement happens when it is the most appropriate (Alg.~\ref{alg:ksumi1}, \textit{Line 8-12}). In the iteration procedure, $D_r$s instead of $C_r$s are maintained and updated. Since this procedure is driven by $\mathcal{I}_m$, it is given as \textit{k}-sums-$\mathcal{I}_m$.

Different from optimization procedure proposed in~\cite{boostkmeans}, \textit{k}-sums aims to minimize the original target function of \textit{k}-means instead of its approximation. The conventional \textit{k}-means clustering is transformed into a pure stochastic minimization process with the target function unchanged. Additionally, our minimization procedure is also essentially different from methods in~\cite{mlpr10:matus,ijcai13:noam}, for which the clustering distortion drops monotonically after each movement. In our method, when moving $x_i$ from $S_w$ to $S_v$, it is the most ``profitable'' act for ``individual'' $x_i$, however this might not be true for other members in $S_v$ and $S_w$. As a result, there will a few bumps in the trend of distortion, while it still shows a general decreasing trend. In contrast, methods in~\cite{boostkmeans,mlpr10:matus,ijcai13:noam} seek for the movement that leads to the decrease of overall distortion in each step. As revealed in the later experiments, the optimization driven by seeking for the better of ``individual interests'' instead of ``general interests'' converges to a better optimum in most of the cases. When $\mathcal{I}_m$ is replaced by $\mathcal{I}_s$ in Alg.~\ref{alg:ksumi1}, it becomes the clustering method driven by $\mathcal{I}_s$, which is given as \textit{k}-sums-$\mathcal{I}_s$.

Fig.~\ref{fig:dist4iter} shows the function value variations after each step (Alg.~\ref{alg:ksumi1}, \textit{Line 6-12}) on four consecutive iterations driven by $\mathcal{I}_m$ and $\mathcal{I}_s$ respectively on a SIFT image feature dataset~\cite{pq}. According to our observation, the distortions from \textit{k}-sums-$\mathcal{I}_m$ decrease steadily as a general trend. However, the function value of Eqn.~\ref{eqn:tkm} may increase in some steps in one round of iteration. This is visible in the zoom-in view of one iteration curve (Fig.~\ref{fig:dist4iter}(b)). This is mainly because \textit{k}-sums-$\mathcal{I}_m$ is driven by ``individual interests'' instead of ``general interests''. Only the distortion associated with a sample is decreased in one movement when driven by $\mathcal{I}_m$ . The movement may lead to the increase of Eqn.~\ref{eqn:tkm} temporarily. However, this invokes other samples (from all clusters) to seek for a better reallocation in the following steps. As a result, the distortion still decreases steadily. The bumps are not observed with \textit{k}-sums-$\mathcal{I}_s$ in Fig.~\ref{fig:dist4iter}(b) since one movement driven by $\mathcal{I}_s$ leads to the steady decrease in both individual distance to a cluster and the overall intra-cluster distances of Eqn.~\ref{eqn:ksumi2}. The function curve of \textit{k}-sums-$\mathcal{I}_s$ that is measured by Eqn.~\ref{eqn:tkm} is shown in Fig.~\ref{fig:dist4iter}(d). The curve shows a general trend of steady decrease. This does indicate two target functions are correlated to some extent. However, they are essentially different given the fact that the decreasing pace in Fig.~\ref{fig:dist4iter}(d) is considerably slower than that of Fig.~\ref{fig:dist4iter}(a).

\section{Extensions and Discussions over \textit{k}-sums}
\label{sec:ext}
\subsection{\textit{k}-means Variants Driven by Optimization}
As presented in Section~\ref{sec:methd}, our modification on \textit{k}-means is simple but fundamental. Theoretically speaking, many \textit{k}-means variants that are built upon the ``egg-chicken'' loop could be optimized following the framework of \textit{k}-sums. In this section, the modification on two popular \textit{k}-means variants is presented. We first consider bisecting \textit{k}-means. Typically, it produces \textit{k} clusters by repeatedly bisecting the intermediate clusters into two~\cite{ml04:zhao}. On each bisecting step, \textit{k}-means is called. As a result, when \textit{k}-sums is adopted in the bisecting step, it becomes bisecting \textit{k}-sums. Moreover, \textit{k}-sums driven by either $\mathcal{I}_s$ or $\mathcal{I}_m$ is feasible. Alg.~\ref{alg:bksum} shows the details of the bisecting \textit{k}-sums.

\begin{algorithm}
    \KwData{Input: matrix $X_{d{\times}n}$, \textit{k}}
    \KwResult{Output: $S_1,{\cdots},S_r,{\cdots},S_k$}
    $S_1 \leftarrow 1{\cdots}n$\;
    Push $S_1$ into a priority queue \textit{Q}\;
    $i \leftarrow 1$\;
    \While {$i < k$}{
    	Pop cluster $S_t$ from \textit{Q}\;
    	Call Alg.~\ref{alg:ksumi1} to cluster $S_t$ into $\{S_{t}^*, S_{i+1}\}$\;
    	Push $S_{t}^*, S_{i+1}$ into queue \textit{Q}\;
    	$i \leftarrow i + 1$\;
    }
   \caption{bisecting \textit{k}-sums}
  \label{alg:bksum}
\end{algorithm}

As shown in Alg.~\ref{alg:bksum}, Alg.~\ref{alg:ksumi1} is called to partition a cluster $S_t$ into two in each step. There could be several ways to decide which cluster $S_t$ to be partitioned. Following the practice in~\cite{ml04:zhao}, cluster with the largest size is selected each time from queue \textit{Q} in our implementation\footnote{In practice, one may choose to split the most loose one.}.

The second \textit{k}-means variant we consider to redefine is Sequential \textit{k}-means, which scans the data only one round and runs online. Given Eqn.~\ref{eqn:tkm} is adopted as the target function for online \textit{k}-means, the update function is revised as
\begin{equation}
D_r=D_r+x_i,
\end{equation}
given that $d(x_i, C_r)=\frac{\parallel{n_r{\cdot}x_i-D_r}\parallel^2}{(n_r+1)^2}$ is the minimum among $k$ clusters. The similar way applies to the case when target function Eqn.~\ref{eqn:ksumi2} is adopted. This revised online clustering method is given as Sequential \textit{k}-sums. Different from conventional Sequential \textit{k}-means, the distance between sample $x_i$ and $C_r$ is calculated assuming that $x_i$ is already joined in $S_r$. The codes of our implementation about \textit{k}-sums and its variants are available at  GitHub\footnote{https://github.com/cc-cyber/k-sums.}.

\subsection{Complexity, Convergence and Optimality Analysis}
\label{sec:cmplx}
It is apparent to see the time complexity of Alg.~\ref{alg:ksumi1} is on the same par as conventional \textit{k}-means. Compared to \textit{k}-means, \textit{k}-sums actually saves up the cost of initial sample-to-centroid assignment, which is equivalent to one round of iteration. In contrast, the time complexity of Hartigan procedure in~\cite{mlpr10:matus,ijcai13:noam} is much higher than it is supposed to be as the optimization is defined on cluster centroids. Unlike conventional \textit{k}-means ``egg-chicken'' loop, the centroid update is a frequent operation in all incremental optimization based methods, namely approaches from~\cite{mlpr10:matus,ijcai13:noam} and \textit{k}-sums. To its worse case, the centroids will be updated \textit{n} times in one round. \textit{k}-sums is computationally more efficient in the sense that it operates on the composite vectors, on which only addition/subtraction operations are involved.

Since Alg.~\ref{alg:ksumi1} could be driven by either $\mathcal{I}_m$ or $\mathcal{I}_s$, the convergence analysis on Alg.~\ref{alg:ksumi1} is divided into two cases. Let's first consider the case as it is driven by $\mathcal{I}_m$. Essentially the iteration is driven by the motivation that $x_i$ seeks for the better allocation such that $d(x_i,C_v)<d(x_i,C_w)$, where $x_i\in S_w$ and is tentatively put into $S_v$. Since $d(x_i, C_v) \ge 0$, there will be a moment for any sample $x_i$ ($d(x_i, C_v) = 0$ to its best) that no movement could take place. At this moment, Alg.~\ref{alg:ksumi1} converges.

When Alg.~\ref{alg:ksumi1} is driven by $\mathcal{I}_s$, it is clear that target function Eqn.\ref{eqn:ksumi2} decreases monotonically after each movement. Given function value Eqn.\ref{eqn:ksumi2} after each movement is $\mathcal{F}^{(t)}$, following inequation series holds.
\begin{equation}
\mathcal{F}^{(1)} > \mathcal{F}^{(2)} > {\cdots} > \mathcal{F}^{(t)} > {\cdots}\geq\mathcal{F}^o,
\label{eqn:convi2}
\end{equation}
where $\mathcal{F}^o$ is the function value as we reach the optimal solution. As a result, the monotonically decreasing function is lower-bounded by $\mathcal{F}^o$. Apparently, it converges.

\begin{figure*}[t]
	\centering
	\subfigure[Measured by $\mathcal{E}_m$]{\includegraphics[width=0.23\linewidth]{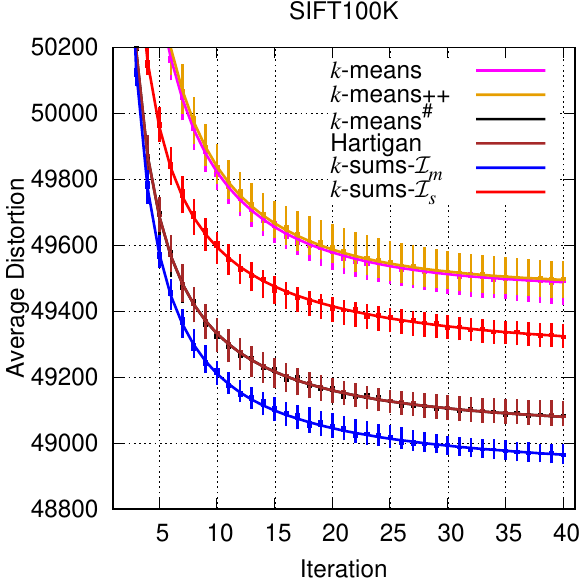}}
	\hspace{0.01in}
	\subfigure[Measured by $\mathcal{E}_s$]{\includegraphics[width=0.238\linewidth]{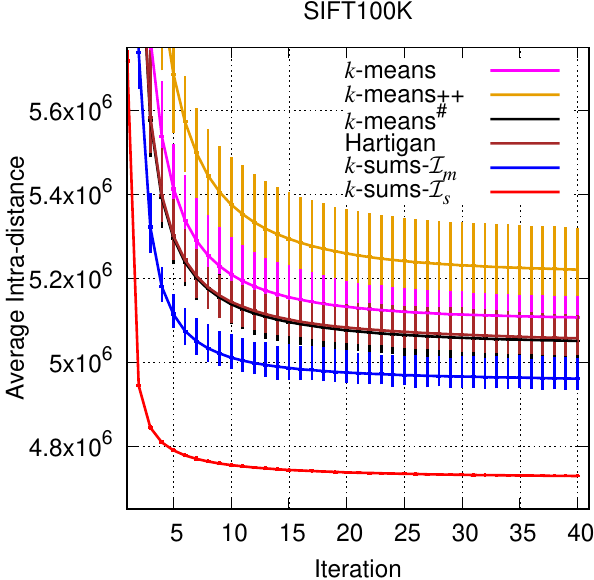}}
	\hspace{0.01in}
	\subfigure[Measured by $\mathcal{E}_m$]{\includegraphics[width=0.245\linewidth]{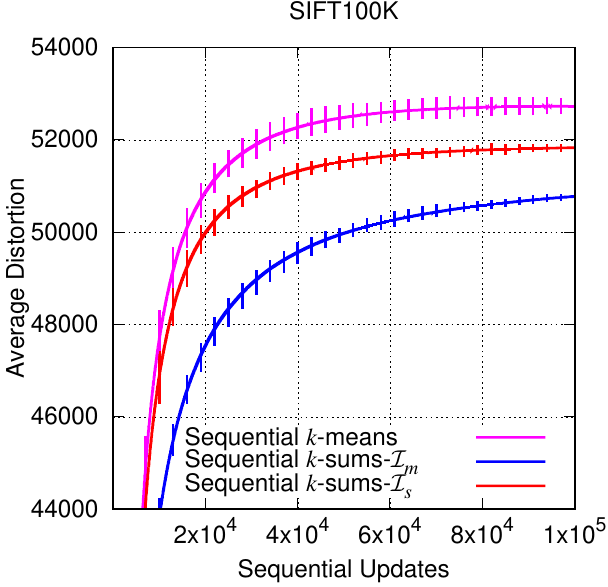}}
	\hspace{0.01in}
	\subfigure[Measured by $\mathcal{E}_s$]{\includegraphics[width=0.247\linewidth]{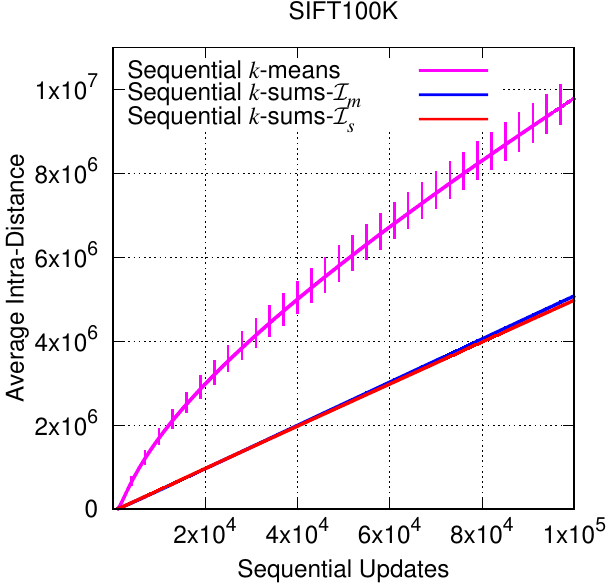}}
	\centering
	\caption{The significance test for \textit{k}-sums-$\mathcal I_m$ and \textit{k}-sums-$\mathcal I_s$. The function values measured by $\mathcal{E}_m$ (figure (a)) and $\mathcal{E}_s$ (figure (b)) are calculated after each iteration. \textit{128} runs are carried out for each method on SIFT100K. The candle chart is plotted based on $\mathcal{E}_m$ and $\mathcal{E}_s$ of 128 runs from each iteration. Notice that all the \textit{k}-means variants minimize Eqn.~\ref{eqn:tkm} except for \textit{k}-sums-$\mathcal{I}_s$. The function values measured by $\mathcal{E}_m$ and $\mathcal{E}_s$ that are produced by Sequential \textit{k}-means and Sequential \textit{k}-sums are shown in figure (c) and (d) respectively.}
	\label{fig:sig}
\end{figure*}

\textit{k}-sums optimization driven either by $\mathcal{I}_m$ or $\mathcal{I}_s$ is greedy. Each optimization step is triggered by the decrease in the distance from an individual sample to its closest centroid (with $\mathcal{I}_m$) or cluster (with $\mathcal{I}_s$). Particularly for \textit{k}-sums-$\mathcal{I}_m$, this is the essential difference as well as the advantage of our method over methods built upon Hartigan procedure~\cite{mlpr10:matus,ijcai13:noam} and \textit{k}-means$^\#$. The minimization in \textit{k}-sums-$\mathcal{I}_m$ is driven by the ``individual interests'' of each sample instead of the ``general interests''  that is regulated by the Hartigan procedure~\cite{mlpr10:matus,ijcai13:noam} . 
The latter imposes implicitly much tighter constraint over the movement of one sample. In these methods, one has to consider the impact to other members from two involved clusters, namely $S_w$ and $S_v$. The ``consensus'' has to be reached among members from two clusters before sample $x_i$ is allowed to move from one to another. In contrast, in \textit{k}-sums-$\mathcal{I}_m$ sample $x_i$ is free to move as long as the new centroid is closer to it than the previous is. It is no need to care about whether this movement is ``beneficial'' to the other members from cluster $S_w$ or $S_v$. Due to the tight constraint, the existing methods~\cite{mlpr10:matus,ijcai13:noam,boostkmeans} tend to be trapped in a local easier than \textit{k}-sums-$\mathcal{I}_m$.

It is possible that other samples in the two involved clusters become further from their centroids after the movement. However, they are therefore invoked to move to other closer clusters under the same rule. As a result, the seemingly ``selfish'' act allows each sample to finally find its closest centroid. Target function Eqn.~\ref{eqn:tkm} is a simple linear summation over distances of each individual to its assigned centroid. The lower of each individual distance leads to the lower overall function value.
%
%
 
Similar as \textit{k}-means, there is no significant change in the structure of the clusters after a few iterations for \textit{k}-sums. Although it turns out to be better than \textit{k}-means and many of its variants, it only reaches a local minimum as \textit{k}-means, \textit{k}-means++ as well as \textit{k}-means$^\#$.


\section{Experiments}
\label{sec:exp}

In this section, the effectiveness of proposed clustering method, namely \textit{k}-sums is studied in comparison to \textit{k}-means and its representative variants. They include \textit{k}-means++~\cite{kpp07}, LVQ~\cite{map01:kohonen}, the method based on Hartigan procedure (given as ``Hartigan'' in the following)~\cite{mlpr10:matus}, \textit{k}-means$^\#$~\cite{boostkmeans}, incremental \textit{k}-means (IKM)~\cite{ml04:zhao}, Sequential \textit{k}-means~\cite{mac67}, Mini-Batch~\cite{mnkm10} and bisecting \textit{k}-means~\cite{ml04:zhao}. For Sequential \textit{k}-means and our redefined Sequential \textit{k}-sums, there is only one single pass over the whole dataset. 

Following the practice in~\cite{ikmn15}, the average distortion (or mean squared error~\cite{pq}) is adopted to evaluate the clustering quality. It is nothing more than the function value of Eqn.~\ref{eqn:tkm} that is averaged by the size of dataset.
The lower the distortion is, the better the clustering quality is. 
\begin{equation}
\mathcal{E}_m = \frac{\sum_{q(x_i)=r}{\parallel C_r -x_i \parallel^2}}{n}
\label{eqn:distor1}
\end{equation}
Similarly, $\mathcal{E}_s$ is introduced to evaluate to what extent target function Eqn.~\ref{eqn:ksumi2} is minimized.
\begin{equation}
\mathcal{E}_s = \frac{\sum_{r=1}^k \sum_{i, j \in S_r \& i < j}{\parallel x_i -x_j \parallel^2}}{n}
\label{eqn:distor2}
\end{equation}

\begin{table}
\begin{center}
\caption{Overview of Datasets}

\begin{tabular}{|l|l|c|}
\hline
Datasets & Scale & Dim. \\ \hline\hline
SIFT100K~\cite{pq} & $1\times10^4$ &  128 \\ \hline
SIFT1M~\cite{pq} & $1\times10^6$   & 128 \\ \hline
GloVe1M~\cite{glove} & $1.1\times10^6$ & 100 \\ \hline
MSD~\cite{ismir12} & $0.99\times10^6$ & 60 \\ \hline
SUSY~\cite{susy14}& $5\times10^6$ & 19 \\ \hline\hline
UMD~\cite{ml04:zhao}& $[878{\sim}9,558]$ & $[2,880{\sim}36,306]$\\ \hline
\end{tabular}

\label{tab:datasets}

\end{center}
\end{table}

Twenty-one datasets are used in the evaluation. The brief information about these datasets is summarized in Tab.~\ref{tab:datasets}. In the first experiment, dataset SIFT100K~\cite{pq} is adopted to perform significance test to confirm that the improvement achieved by our approach is not by random. In the second experiment, \textit{k}-sums is tested on four large-scale datasets. The types of data range from image local features (SIFT1M)~\cite{pq}, vectorized text word features (GloVe1M)~\cite{glove}, to audio features (MSD)~\cite{ismir12} and event descriptions (SUSY)~\cite{susy14}. In the last experiment, \textit{15} document datasets (UMD)~\cite{ml04:zhao} are adopted. The documents are represented with TF/IDF model and are $\textit{l}_2$-normalized. On this document clustering task, the performance is evaluated by entropy~\cite{ml04:zhao}.

\begin{equation}
Entropy = \sum_{r=1}^k\frac{n_r}{n}\frac{1}{\log{c}}*\sum_{i=1}^c{\frac{n_r^i}{n_r}*\log{\frac{n_r^i}{n_r}}},
\label{eqn:entropy}
\end{equation}
where \textit{c} is the number of classes in the ground-truth, and $n_r^i$ is the size of intersection between class \textit{i} and cluster $S_r$. The entropies obtained from \textit{15} document datasets are averaged for each considered method.

\subsection{Significance Test}
The initialization on \textit{k}-means clustering is based on either random seeding or random label assignment. Moreover, the optimization is a stochastic procedure for the methods such as IKM, \textit{k}-means$^\#$, Hartigan, and \textit{k}-sums. For these two reasons, the clustering results from \textit{k}-means and its variants vary from one run to another. The first experiment investigates the general performance trends of \textit{k}-sums-$\mathcal{I}_m$ and \textit{k}-sums-$\mathcal{I}_s$ and the variations across different runs. The experiment is conducted on SIFT100K. For each considered method, \textit{128} runs are undertaken. The cluster number \textit{k} is fixed to \textit{1,024}. $\mathcal{E}_m$ and $\mathcal{E}_s$ are calculated after one iteration.

The candle charts for four methods from $\mathcal{E}_m$ and $\mathcal{E}_s$ are shown in Fig.~\ref{fig:sig}(a) and Fig.~\ref{fig:sig}(b) respectively. The trend curves produced by Sequential \textit{k}-means and Sequential \textit{k}-sums with respect to $\mathcal{E}_m$ and $\mathcal{E}_s$ are shown in Fig.~\ref{fig:sig}(c) and Fig.~\ref{fig:sig}(d). As shown from the figure, \textit{k}-sums-$\mathcal{I}_m$ and \textit{k}-sums-$\mathcal{I}_s$ achieve the lowest function score with respect to their target functions after \textit{3} iterations. The performance gap between our methods and the rest is much more significant than the possible variations between different runs. As \textit{k}-sums-$\mathcal{I}_s$ is the only method that aims to minimize target function Eqn.~\ref{eqn:ksumi2}, a wide performance gap is observed in Fig.~\ref{fig:sig}(b). The performance from Hartigan nearly overlaps with that of \textit{k}-means$^\#$. Although \textit{k}-means$^\#$ addresses \textit{k}-means clustering as a maximization problem,  it behaves similarly as Hartigan~\cite{mlpr10:matus} as both of them incrementally optimize the \textit{k}-means target function in a monotonic manner. The performance gap between \textit{k}-means and \textit{k}-means++ is nearly invisible from Fig.~\ref{fig:sig}(a). This indicates the improvement from  seeding scheme is limited. In terms of online \textit{k}-means, all the curves given by $\mathcal{E}_m$ and $\mathcal{E}_s$ rise up as more and more samples join in. This is because the overall function values of $\mathcal{E}_m$ and $\mathcal{E}_s$ increase as more samples are incorporated in the equation. As shown in Fig.~\ref{fig:sig}(c) and Fig.~\ref{fig:sig}(d), Sequential \textit{k}-sums show the lowest function value in each iteration with respect to the corresponding target function. Moreover, they demonstrate a much narrower variation range than that of Sequential \textit{k}-means.

\subsection{Quality Evaluation on Various Data Types}
\begin{figure}[t]
	\centering
	\subfigure[]{\includegraphics[width=0.473\linewidth]{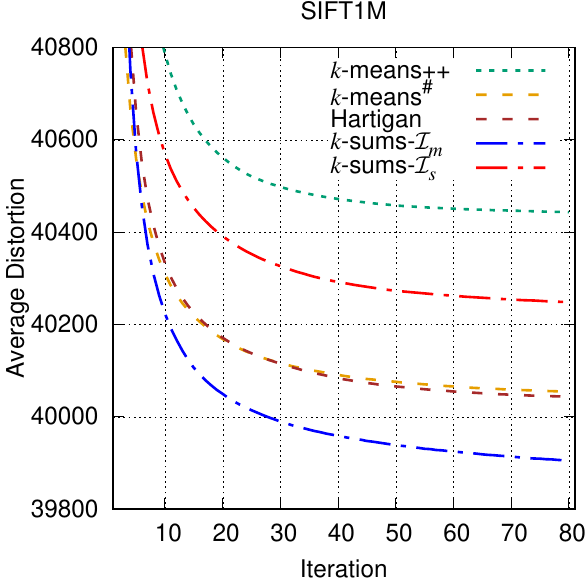}}
	\hspace{0.01in}
	\subfigure[]{\includegraphics[width=0.465\linewidth]{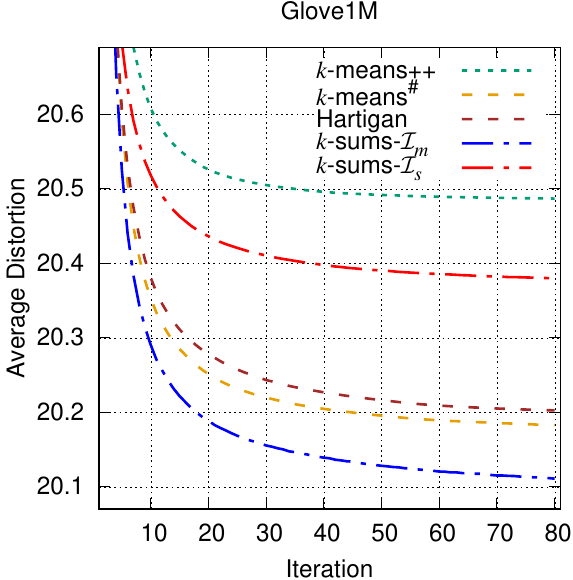}}
	\subfigure[]{\includegraphics[width=0.45\linewidth]{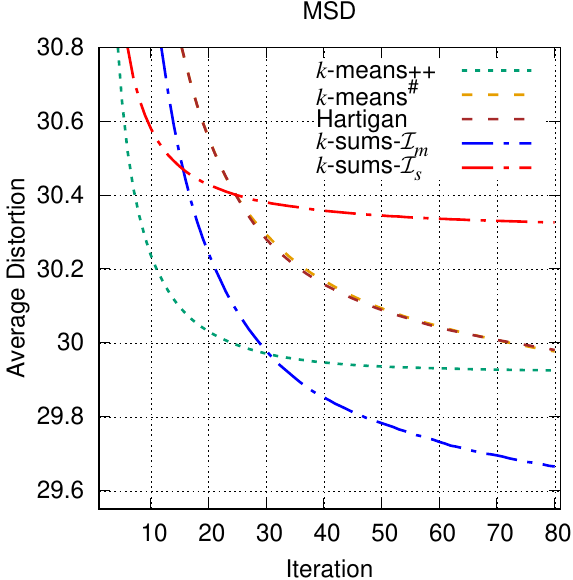}}
	\hspace{0.01in}
	\subfigure[]{\includegraphics[width=0.466\linewidth]{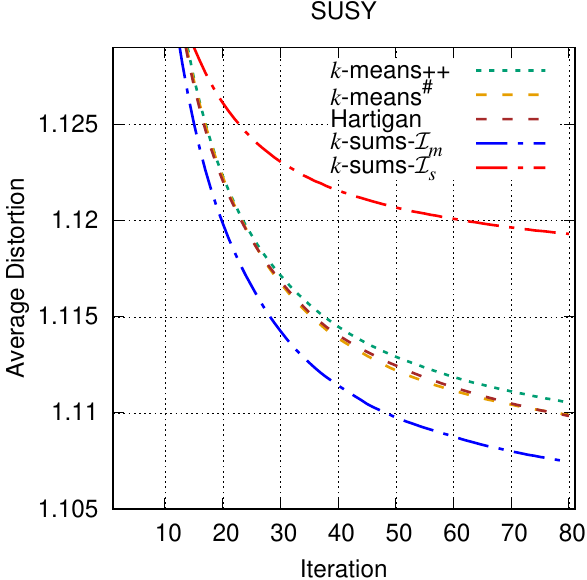}}
	\centering
	\caption{The general trend of function values measured by $\mathcal{E}_m$. }
	\label{fig:distI1}
\end{figure}

\begin{figure}[t]
	\centering
	\subfigure[]{\includegraphics[width=0.48\linewidth]{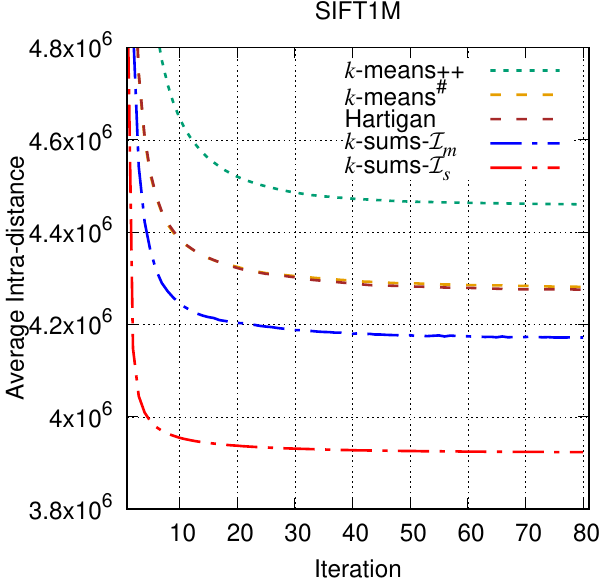}}
	\hspace{0.01in}
	\subfigure[]{\includegraphics[width=0.46\linewidth]{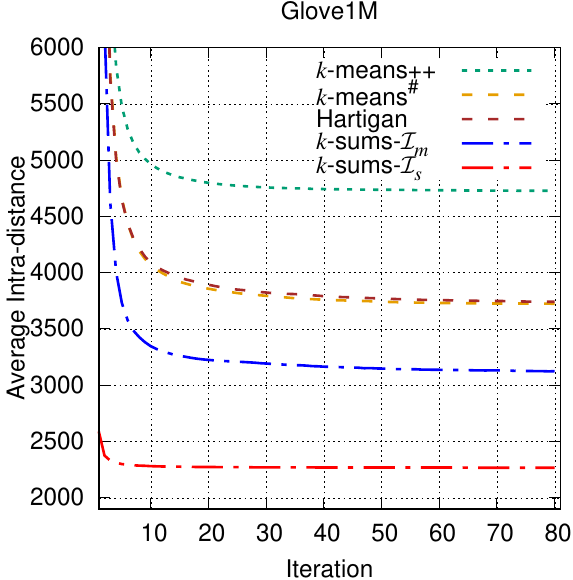}}
	\subfigure[]{\includegraphics[width=0.475\linewidth]{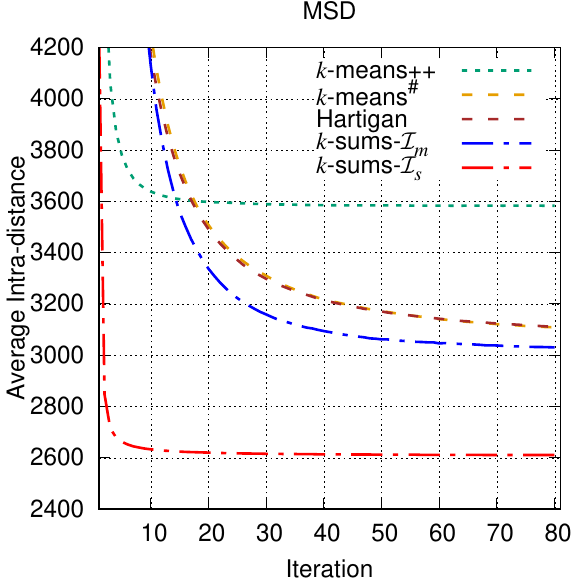}}
	\hspace{0.01in}
	\subfigure[]{\includegraphics[width=0.466\linewidth]{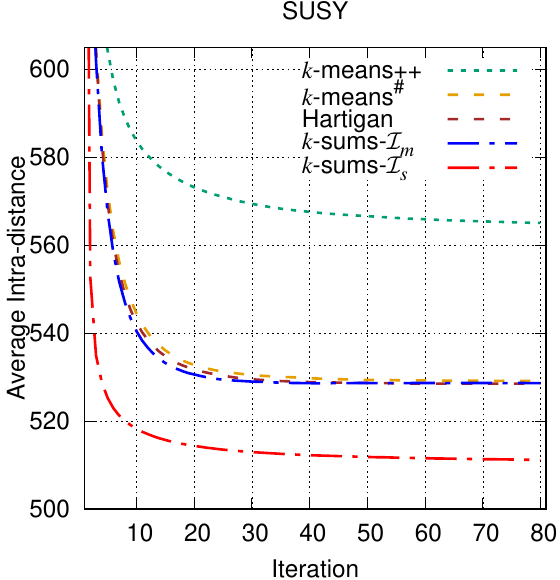}}
	\centering
	\caption{The general trend of function values measured by $\mathcal{E}_s$. }
	\label{fig:distI2}
\end{figure}

In the second experiment, four large-scale datasets of various data types are adopted in the evaluation. They are SIFT1M, GloVe1M, MSD and SUSY. The general trends of $\mathcal{E}_m$ and $\mathcal{E}_s$ from \textit{k}-means$^\#$, Hartigan, \textit{k}-means++, \textit{k}-sums-$\mathcal{I}_m$ and \textit{k}-sums-$\mathcal{I}_s$ are studied on these datasets. Since \textit{k}-means++ usually shows better clustering quality than \textit{k}-means and many other variants, it is treated as the comparison baseline. \textit{k} is fixed to \textit{10,000} for all the methods on each dataset. According to the previous experiment, the performance gap between the  methods is more significant than the possible variations between different runs. It is therefore valid to only show the distortion curve of one run. The curves from $\mathcal{E}_m$ and $\mathcal{E}_s$ are shown in Fig.~\ref{fig:distI1} and Fig.~\ref{fig:distI2} respectively.  

As shown in the figures, \textit{k}-sums remains the best method with respect to two evaluation criterion, which is consistent with the previous observations. The function values from \textit{k}-sums decrease at a much faster pace than the other three methods. Moreover, the performance gap gets wider as the number of iterations grows for \textit{k}-sums-$\mathcal{I}_m$. This basically indicates that \textit{k}-sums less likely gets trapped in a local minimum when driven by $\mathcal{I}_m$. In contrast, the rankings of cluster quality from \textit{k}-means++ and \textit{k}-means$^\#$ vary across different datasets. Generally the performance becomes saturated within \textit{30} iterations for both of them. Similar as the previous observation, the performance trend from \textit{k}-means$^\#$  and Hartigan remains similar.  \textit{k}-sums-$\mathcal{I}_s$ shows poorer performance than the others in Fig.~\ref{fig:distI1}. However, it converges quickly to a much better local optimum than the rest when measured by $\mathcal{E}_s$ (shown in Fig.~\ref{fig:distI2}). It is the only method that is designed to minimize target function Eqn.~\ref{eqn:ksumi2}. This indicates two target functions considered in the paper are correlated yet still essentially different.


\subsection{Document Clustering}
In the third experiment, the performance of our methods is studied on the classic document clustering task. Fifteen datasets from UMD are adopted. In the experiments, \textit{k}-means and the other five variants are considered. \textit{Cosine} distance is adopted for all the methods. For the methods such as \textit{k}-means, \textit{k}-means++, \textit{k}-means$^\#$ and \textit{k}-sums, they could be undertaken in a bisecting manner, namely in the way of Alg.~\ref{alg:bksum}. As a result, the performance under the bisecting strategy for these methods is also reported. For each method, \textit{k} is set to \textit{5}, \textit{10}, \textit{15} and \textit{20} on each dataset. Following the practice in~\cite{ml04:zhao}, the clustering result of one method is selected from \textit{10} runs with the lowest $\mathcal{E}_m$ or $\mathcal{E}_s$ for \textit{k}-sums-$\mathcal{I}_s$. The average entropies of each method with both the direct \textit{k}-way and the bisecting clustering are reported on Tab.~\ref{tab:umn}(a) and Tab.~\ref{tab:umn}(b) respectively.
\begin{table}[!t]
	\caption{Clustering performance on UMD \textit{15} document datasets}
	\label{tab:umn}
	\centering
	\subtable[Clustering performance by direct \textit{k}-way]{
			\begin{tabular}{|l|c|c|c|c|}
				\hline
				& \textit{k}=5   & \textit{k}=10  & \textit{k}=15  & \textit{k}=20 \\
				\hline
				\hline
				\textit{k}-means~\cite{km82} & 0.539 & 0.443 & 0.402 & 0.387 \\
				\hline
				\textit{k}-means++~\cite{kpp07} & 0.550 & 0.441 & 0.403 & 0.389 \\
				\hline
				Mini-Batch~\cite{mnkm10} & 0.585 & 0.488 & 0.469 & 0.475 \\
				\hline
				LVQ~\cite{map01:kohonen}   & 0.800 & 0.761 & 0.681 & 0.674 \\
				\hline
				\textit{k}-means$^\#$~\cite{boostkmeans} & 0.552 & 0.442 & 0.388 & 0.368 \\
				\hline
				Hartigan~\cite{mlpr10:matus}   & 0.451 & 0.358 & 0.331 & \textbf{0.307}  \\
				\hline
				IKM~\cite{ml04:zhao}   & 0.465 & 0.401 & 0.366 & 0.358 \\
				\hline
				\hline
				\textit{k}-sums-$\mathcal I_m$ & 0.452 & 0.362 & 0.330 & 0.312 \\
				\hline
				\textit{k}-sums-$\mathcal I_s$ & \textbf{0.445} & \textbf{0.357} & \textbf{0.325} & 0.308 \\
				\hline
			\end{tabular}
			\label{tab:directkway}
		}
	\subtable[Clustering performance by bisecting]{          
			\begin{tabular}{|l|c|c|c|c|}
				\hline
				& \textit{k}=5   & \textit{k}=10  & \textit{k}=15  & \textit{k}=20 \\
				\hline
				\hline
				\textit{k}-means~\cite{km82} & 0.532 & 0.438 & 0.410 & 0.373 \\
				\hline
				\textit{k}-means++~\cite{kpp07} & 0.507 & 0.422 & 0.400 & 0.379 \\
				\hline
				\textit{k}-means$^\#$~\cite{boostkmeans} & 0.514 & 0.388 & 0.353 & 0.329 \\
				\hline
				IKM~\cite{ml04:zhao}   & 0.465 & 0.390 & 0.353 & 0.330 \\
				\hline
				\hline
				\textit{k}-sums-$\mathcal I_m$ & \textbf{0.449} & \textbf{0.367} & \textbf{0.335} & \textbf{0.311} \\
				\hline
				\textit{k}-sums-$\mathcal I_s$ & 0.494 & 0.408 & 0.359 & 0.345 \\
				\hline
			\end{tabular}
			\label{tab:bisecting}
	}  
\end{table}

As shown on the tables, \textit{k}-sums driven by $\mathcal I_m$ and $\mathcal I_s$ outperform other methods considerably on the direct \textit{k}-way case. On the bisecting case, \textit{k}-sums-$\mathcal I_m$ still shows the best results, while \textit{k}-sums-$\mathcal I_s$ shows similar performance as \textit{k}-means$^\#$. \textit{k}-sums-$\mathcal I_s$ shows relatively poor performance because it converges quickly and therefore is unable to reach a better local optimum in the bisecting case. IKM is the only method that shows close performance with \textit{k}-sums. Unfortunately, it only works under \textit{Cosine} distance~\cite{boostkmeans, ml04:zhao}.  \textit{k}-means$^\#$ and Hartigan perform similarly as they essentially optimize the target function in the similar manner. As explained in Section~\ref{sec:cmplx}, both of them tend to be trapped in a local optimum easier than \textit{k}-sums due to the tight constraint over the sample reallocation.

\section{Conclusion}
\label{sec:conc}
In this paper, the simple ``egg-chicken'' loop in \textit{k}-means has been modified to an even simpler stochastic optimization procedure. Different from \textit{k}-means and many of its variants, the distortion minimization is driven by seeking for the better reallocation of each individual sample. The clusters are updated as soon as the reallocation of one sample leads to the lower distortion that is associated with the sample. A family of \textit{k}-means variants are redefined under this optimization framework and show considerably better clustering quality. Moreover, another target function is proposed to handle the case that cluster centroid/mode cannot be defined. It is then solved under the same optimization procedure. To generalize this new clustering model to the generic metric space is our future research direction.

{\small
\bibliographystyle{ieee_fullname}
\bibliography{wlzhao.bib}

\begin{thebibliography}{10}\itemsep=-1pt

\bibitem{kpp07}
David Arthur and Sergei Vassilvitskii.
\newblock k-means++: The advantages of careful seeding.
\newblock In {\em Proceedings of the Eighteenth Annual ACM-SIAM Symposium on
  Discrete Algorithms}, pages 1027--1035, 2007.

\bibitem{ikmn15}
Yannis Avrithis, Yannis Kalantidis, Evangelos Anagnostopoulos, and Ioannis~Z.
  Emiris.
\newblock Web-scale image clustering revisited.
\newblock In {\em ICCV}, pages 1502--1510, Dec. 2015.

\bibitem{nips16bachem}
Olivier Bachem, Mario Lucic, Hamed Hassani, and Andreas Krause.
\newblock Fast and provably good seedings for k-means.
\newblock In {\em Advances in Neural Information Processing Systems 29}, pages
  55--63, 2016.

\bibitem{kpp12}
Bahman Bahmani, Benjamin Moseley, Andrea Vattani, Ravi Kumar, and Sergei
  Vassilvitskii.
\newblock Scalable k-means++.
\newblock {\em In Proceedings of the VLDB Endowment}, 5(7):622--633, 2012.

\bibitem{susy14}
Pierre Baldi, Przemysław Sadowski, and D. Whiteson.
\newblock Searching for exotic particles in high-energy physics with deep
  learning.
\newblock {\em Nature communications}, 5:4308, Jul. 2014.

\bibitem{ikm}
Andrei Broder, Lluis Garcia-Pueyo, Vanja Josifovski, Sergei Vassilvitskii, and
  Srihari Venkatesan.
\newblock Scalable k-means by ranked retrieval.
\newblock In {\em Proceedings of the 7th ACM international conference on Web
  search and data mining}, pages 233--242, 2014.

\bibitem{wsdm14}
Andrei Broder, Lluis Garcia-Pueyo, Vanja Josifovski, Sergei Vassilvitskii, and
  Srihari Venkatesan.
\newblock Scalable k-means by ranked retrieval.
\newblock In {\em Proceedings of the 7th ACM international conference on Web
  search and data mining}, pages 233--242, Feb. 2014.

\bibitem{meansift}
Yizong Cheng.
\newblock Mean shift, mode seeking, and clustering.
\newblock {\em Trans. PAMI}, 17(8):790--799, Aug. 1995.

\bibitem{dbscan}
Martin Ester, Hans peter Kriegel, Jörg Sander, and Xiaowei Xu.
\newblock A density-based algorithm for discovering clusters in large spatial
  databases with noise.
\newblock In {\em {\sc IEEE} Transactions on Knowledge and Data Engineering},
  pages 226--231, 1996.

\bibitem{icdm04}
A. Goswami, Ruoming Jin, and G. Agrawal.
\newblock Fast and exact out-of-core k-means clustering.
\newblock In {\em Fourth IEEE International Conference on Data Mining}, pages
  83--90, Nov. 2004.

\bibitem{cluster75:hartigan}
John~A. Hartigan.
\newblock {\em Clustering Algorithms (Probability \& Mathematical Statistics)}.
\newblock John Wiley \& Sons Inc., 1975.

\bibitem{jain88}
Anil~K. Jain and Richard~C. Dubes.
\newblock {\em Algorithms for Clustering Data}.
\newblock 1988.

\bibitem{Jain99}
A.~K. Jain, M.~N. Murty, and P.~J. Flynn.
\newblock Data clustering: A review.
\newblock {\em ACM Computing Surveys}, 31(3):264--323, Sep. 1999.

\bibitem{pq}
Herv\'e J\'egou, Matthijs Douze, and Cordelia Schmid.
\newblock Product quantization for nearest neighbor search.
\newblock {\em Trans. PAMI}, 33(1):117--128, Jan. 2011.

\bibitem{JPDSPS11}
Herv\'e J\'egou, Florent Perronnin, Matthijs Douze, Jorge S\'anchez, Patrick
  P\'erez, and Cordelia Schmid.
\newblock Aggregating local descriptors into compact codes.
\newblock {\em Trans. PAMI}, 34(9):1704--1716, Sep. 2012.

\bibitem{pam87:kaufman}
Leonard Kaufman and Peter~J. Rousseeuw.
\newblock Clustering by means of medoids.
\newblock {\em Reports of the Faculty of Mathematics and Informatics},
  87:405--416, 1987.

\bibitem{map01:kohonen}
T. Kohonen, M.~R. Schroeder, and T.~S. Huang, editors.
\newblock {\em Self-Organizing Maps}.
\newblock Springer-Verlag New York, Inc., Secaucus, NJ, USA, 3rd edition, 2001.

\bibitem{km82}
Stuart~P. Lloyd.
\newblock Least squares quantization in {PCM}.
\newblock {\em IEEE Trans. Information Theory}, 28:129--137, Mar. 1982.

\bibitem{kmeans}
MacQueen, James, et~al.
\newblock Some methods for classification and analysis of multivariate
  observations.
\newblock In {\em Proceedings of the fifth Berkeley symposium on mathematical
  statistics and probability}, volume~1, pages 281--297, 1967.

\bibitem{mac67}
J. MacQueen.
\newblock Some methods for classification and analysis of multivariate
  observations.
\newblock In {\em Proceedings of 5th Berkeley Symposium on Mathematical
  Statistics and Probability}, pages 281--297, 1967.

\bibitem{rankorder}
C. Otto, D. Wang, and A. Jain.
\newblock Clustering millions of faces by identity.
\newblock {\em Trans. PAMI}, pages 1--14, Mar. 2017.

\bibitem{pelleg99}
Dan Pelleg and Andrew Moore.
\newblock Accelerating exact k-means algorithms with geometric reasoning.
\newblock In {\em Proceedings of the Fifth ACM SIGKDD International Conference
  on Knowledge Discovery and Data Mining}, pages 277--281, Aug. 1999.

\bibitem{glove}
Jeffrey Pennington, Richard Socher, and Christopher~D. Manning.
\newblock Glove: Global vectors for word representation.
\newblock In {\em Empirical Methods in Natural Language Processing (EMNLP)},
  pages 1532--1543, 2014.

\bibitem{science14}
Alex Rodriguez and Alessandro Laio.
\newblock Clustering by fast search and find of density peaks.
\newblock {\em Science}, 344(6191):1492--1496, 2014.

\bibitem{ismir12}
Alexander Schindler, Rudolf Mayer, and Andreas Rauber.
\newblock Facilitating comprehensive benchmarking experiments on the million
  song dataset.
\newblock In {\em In Proceedings of the 13th International Society for Music
  Information Retrieval Conference}, pages 469--474, Dec. 2012.

\bibitem{mnkm10}
D. Sculley.
\newblock Web-scale k-means clustering.
\newblock In {\em Proceedings of the 19th international conference on World
  wide web}, pages 1177--1178, 2010.

\bibitem{ijcai13:noam}
Noam Slonim, Ehud Aharoni, and Koby Crammer.
\newblock Hartigan's k-means versus lloyd's k-means: is it time for a change?
\newblock In {\em Proceedings of the Twenty-Third international joint
  conference on Artificial Intelligence}, pages 1677--1684, 2013.

\bibitem{mlpr10:matus}
Matus Telgarsky and Andrea Vattani.
\newblock Hartigan's method: k-means clustering without voronoi.
\newblock In {\em Proceedings of the Thirteenth International Conference on
  Artificial Intelligence and Statistics}, pages 820--827, 2010.

\bibitem{spectral}
Ulrike von Luxburg.
\newblock A tutorial on spectral clustering.
\newblock {\em Statistics and Computin}, 17(4):395--416, Aug. 2007.

\bibitem{top10}
Xindong Wu, Vipin Kumar, J.~Ross Quinlan, Joydeep Ghosh, Qiang Yang, Hiroshi
  Motoda, Geoffrey~J. McLachlan, Angus Ng, Bing Liu, Philip~S. Yu, Zhi-Hua
  Zhou, Michael Steinbach, David~J. Hand, and Dan Steinberg.
\newblock Top 10 algorithms in data mining.
\newblock {\em Knowledge and Information System}, 14(1):1--37, Dec. 2007.

\bibitem{boostkmeans}
Wan-Lei Zhao, Cheng-Hao Deng, and Chong-Wah Ngo.
\newblock k-means: a revisit.
\newblock {\em Neurocomputing}, 291:195--206, 2018.

\bibitem{ml04:zhao}
Ying Zhao and Geoge Karypis.
\newblock Empirical and theoretical comparisons of selected criterion functions
  for document clustering.
\newblock {\em Machine Learning}, 55:311--331, Jun. 2004.

\bibitem{kddzhao05}
Ying Zhao and George Karypis.
\newblock Hierarchical clustering algorithms for document datasets.
\newblock {\em Data Mining and Knowledge Discovery}, 10(2):141--168, Mar. 2005.

\end{thebibliography}
}

\end{document}